# Improved Human Emotion Recognition Using Symmetry of Facial Key Points with Dihedral Group

Mehdi Ghayoumi
Artificial Intelligence Lab
Computer Science Department
Kent State University
Ohio-USA
mghayoum@kent.edu

Arvind K. Bansal
Artificial Intelligence Lab
Computer Science Department
Kent State University
Ohio-USA
akbansal@kent.edu

*Abstract:* this article describes how to deploy dihedral group theory to detect Facial Key Points (FKP) symmetry to recognize emotions. The method can be applied in many other areas which those have the same data texture.

## 1 Introduction

These days, Human Computer Interaction (HCI) is one of the areas in computer science which is very crowded and can it be applied to many purposes and applications. There are many types of information which can help to the human communication with the machine and facial expression provides a large measure of these data. There are many researchers in this area, who they are working for many purposes, especially for emotion analysis [31], [32].

If we take a look, we can find that almost all human activities are affected by emotion and it directly or indirectly makes huge differences in all the task's performance. Emotion analysis, especially, can be used in machines and when the machine has it, then it can have better interaction with humans and other machines. But besides all researches in this area, we still have some issues. For example, the accuracy and real time analysis are still some challenges in this field [23].

One of the main goals in computer science, is doing less designing, computing and testing, which deploy less computation resources and then gives us better performance and one of the topics which recently has received especial researcher's attention, is finding the symmetry in objects. Symmetry can be found in many objects in the nature and can be used for many purposes. Based on the human face structure, it usually can be found in facial images and then using such this technique can make many types of performances better [1, 2 and 3]

Automatic symmetry detection on 2D or 3D images has been an active research field, in the recent decades [4, 5 10 and 11]. Through these researches, most of the time, the best style of receiving an accurate and robust result is a mathematical method [22], [24] [14], [12] and the Group Theory is one of these mathematical concepts.

One type of the Group Theory, which regularly deploy in the symmetry applications is dihedral groups which are created from symmetries of regular polygons [13]. It has many applications in chemical, geometry and recently computer science [25], [26] [27] and [28]. In our last research the method for applying dihedral groups for real emotion analysis presented [42] and in this research, we'll study this symmetry group, to give some robust strategy for flipping and rotating operation with specific degree for finding symmetry in facial image. It can be used for training some learning methods such as neural networks or convolution neural network.

The rest of the article is as follows: Section 2 presents a brief explanation of the human emotion analysis. Section 3 explains the symmetry in the face. Section 4 describes the Dihedral Groups fundamental with an example for rectangles. Section 5 describes how to use the dihedral group for facial symmetry detections and the last section concludes the paper and explain the future works.

## 2 Human Emotion Analysis

Analyzing human emotion make the Human-Machine Interaction (HMI) and Machine-Machine Interaction (MMI) more efficient. It can be done in two completely different processes which are connected to each other:

1. Emotion Recognition (ER) and
2. Emotion Generation (EG).

ER can be mined from some information such as visual and audio data which are more important in the human emotion definition. The most data from video or image can come up from facial images which have a lot of information and facial expression can be recognized by 3 methods:

a) Geometric feature-based method [43],
b) Appearance-based method [23], and
c) Hybrid-based method [44].

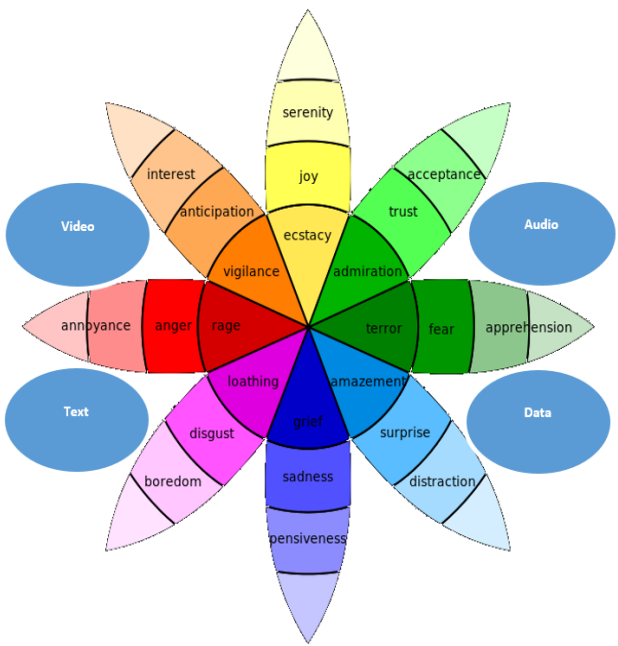

**Fig 1**. Some human emotion states

Figure 1 shows a general view for some basic and secondary human emotions which all can be extracted from video, audio, text and some other data such as silence and its duration. There are also some relations between these data which can present better and more accurate emotion state, especially when we are looking for mixed emotion that are more complicated [29].

## 3 Symmetry in Face

The facial characteristics are determined by the skull shape [16], [15]. Take a photo of the whole face from front side and folds in half and flip the picture complete into side by side by the mirrored version of the image. Compare the photo of the original face with the revised version of the mirrored versions [3], [4], [6]. Left side face or right one, which one is closer to the original. These are the questions which are related to many parameters. And if you chose the face with emotional states, which one show the emotional state more? Based on the researches the left face displays emotions more intensely than the right face [7], [8] and [9]. So there are some ideas here, at the first three are some symmetries and asymmetries which can be extracted from facial image and for emotion analysis purpose, working on the left side is more effective. Figure 2 shows the 2 faces which are constructed of 2 half symmetry parts of the original face.

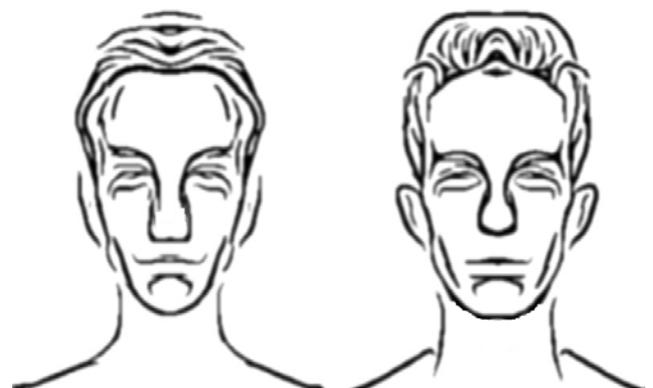

**Fig. 2**. **a**. From left side **b**. Form right side

There are many facial key points which are important in emotion analysis through the face. There are following features pairs (left and right) for the analysis of symmetry:

• Eyebrows (3 points),
• Eyes (4 points),
• Lips side (3 points),
• Middle of Lips (4 points).

All of these points are in left and right sides and then in total we have 24 points, 10 points on each side and 4 points in the middle of the lips.

Figure 3, shows facial key points. The feature points which are colored red, blue and green representing respectively a stable, active or passive state during an expression. Facial symmetry is measured through changing asymmetry of the face [17, 18, 19, 20, 21, 30, 39, 38, 37, 36 and 40] and the symmetry contains rotation and reflection which both can be occur in the facial features.

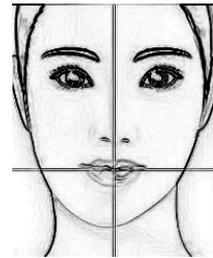

**Fig. 4**. Face with normal central axis

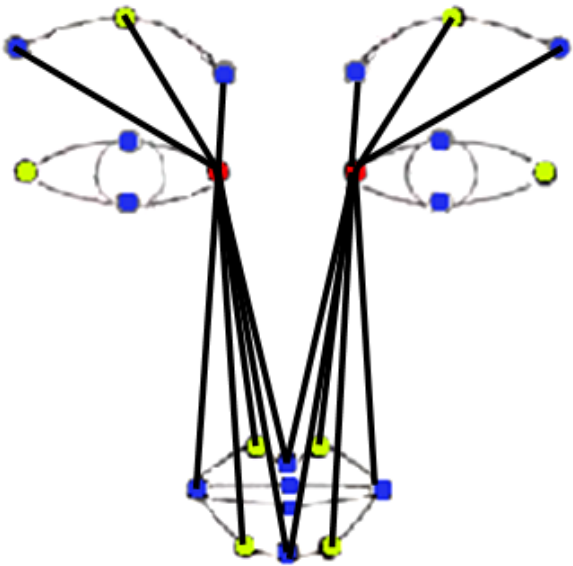

**Fig 3.** Face model showing the feature points [red=stable, blue=active, and the remainders are passive]

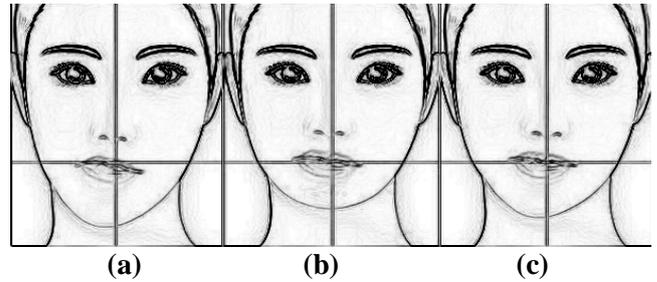

(a) (b) (c)

**Fig. 5 a.** Central axis with protruded chain, **b.** Central axis with protruded mouth and a little chain and the **c.** Asymmetry on the chain.

On the other hand, the asymmetric facial expression can be categorized in two types:

**a)** Asymmetry in movement and
**b)** Structural asymmetry in the face.

Both of these methods can be applied for finding out the asymmetry in the face. Movement is generally the primary source of finding out the asymmetry in the emotional facial expression state and it can be measured by finding the pixel value change through the time. The structural asymmetry of emotional expression facial expression state has main implications for evolutionary interpretations of facial expressions state. Figure 4 and 5 are some examples of normal and asymmetry states on the face. Figure 4 shows a normal image with 2 central axis and figure 5.a, b, and c shows some asymmetry on the face on chain and moth (lips). Finding the asymmetry even can help us to extract some information about human emotion. For example, in the last research, we discussed about real emotion which can be extracted from asymmetry on facial image data [42].

We have 6 basic emotion states which for each one some action units should put on to face as shown in Table 1. By detecting correct symmetry points, we can predict other correspond points and it can help us for emotions analyzing. It is not usable for all emotion stats, (especially mixed emotions) but for some purpose when the faces are occluded can be used. Table 2 shows the AUs which relate to the 6 basic emotion states and Table 3 shows the list of Active Action Units (AAU). We divide the AUs to the active (1, 2,4,12,15,16,20 and 23) and passive (5, 6, 7, 9 and 26) and then refine Table 2 & 3, in Table 4 to realize the symmetry for emotion analysis.

**Table 1.** List of Action Units numbers and their Action Descriptors

| AU Number | FACS |
|---|---|
| 1 | Inner Brow Raiser |
| 2 | Outer Brow Raiser |
| 4 | Brow Lowerer |
| 5 | Upper Lid Raiser |
| 6 | Cheek Raiser |
| 7 | Lid Tightener |
| 9 | Nose Wrinkler |
| 12 | Lip Corner Puller |
| 15 | Lip Corner Depressor |
| 16 | Lower Lip Depressor |
| 20 | Lip Stretcher |
| 23 | Lip Tightener |
| 26 | Jaw Drop |

**Table 2.** Action Units for Basic Emotions

| Emotions | Action Units |
|---|---|
| Happiness | 6+12 |
| Sadness | 1+4+15 |
| Surprise | 1+2+5+26 |
| Fear | 1+2+4+5+7+20+26 |
| Anger | 4+5+7+23 |
| Disgust | 9+15+16 |

**Table 3.** List of Active Action Unit

| FACS | AU Number |
|---|---|
| Inner Brow Raiser | 1 |
| Outer Brow Raiser | 2 |
| Brow Lowerer | 4 |
| Lip Corner Puller | 12 |
| Lip Corner Depressor | 15 |
| Lower Lip Depressor | 16 |
| Lip Stretcher | 20 |
| Lip Tightener | 23 |

**Table 4.** Refined Action Units for Basic Emotions

| Emotions | Action Units |
|---|---|
| Happiness | 12 |
| Sadness | 1+4+15 |
| Surprise | 1+2 |
| Fear | 1+2+4+20 |
| Anger | 4+23 |
| Disgust | 15+16 |

# 4 Dihedral Groups

The dihedral group has two major operations: rotation and reflection. The regular polygon with **n** sides is a dihedral group and can be present by $D_n$ and has **2n** elements as:

$$\{e, r, r^2, \ldots, r^{n-1}, s, sr, sr^2, \ldots, sr^{n-1}\} \quad (1)$$

Where **e** is the identity element in $D_n$. We can write $D_n$ as:

$$D_n = \{s_j r^k : 0 \le k \le n-1, 0 \le j \le 1\} \quad (2)$$

Which has the following properties:

$$r^n = 1, sr^k s = r^{-k}, (sr^k)^2 = e, \text{ for all } 0 \le k \le n-1 \quad (3)$$

And the composition of two elements of the $D_n$ is given by:

$$r^i r^j = r^{i+j},\ r^i sr^j = sr^{j-i},\ sr^i r^j = sr^{i+j},\ sr^i sr^j = r^{j-i} \quad (4)$$

Here, as an example, we explain $D_4$, because the face can be inscribed in a rectangle in the simplest structure which cover most data besides keeping lets computation. We will consider the rotations and reflections of a square denoted as $D_4$ represented visually as four counterclockwise rotations of a square. We work on brows, eyes, and the lips. For example, in the figure 4, number 1 is regarded as the right brow facial point and the number 2 is left brow and eye center points and 3 and 4 are right and left lip corner points in the facial key points respectively and then We can take on the dihedral properties and matrices for rotation and reflection matrices as follows:

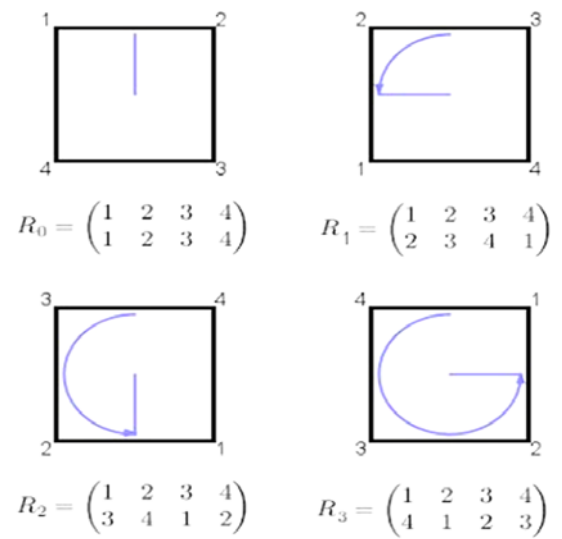

**Fig. 4.** Four counterclockwise rotations of a square and their transformation matrixes for facial key point detection.

In the Figure .4, **R$_i$** for **i** ∈ {0, 1, 2, 3} are as follows:

- **R$_0$** is the identity (0 or 360 rotation),
- **R$_1$** is a rotation of 90,
- **R$_2$** is the rotation of 180 and
- **R$_3$** is the rotation of 270.

And the following four reflections, vertical, horizontal, and two diagonal reflections and their matrices. In the future works these matrices will deploy as filter for image convolving for applying on facial image and finding out the symmetry points especially when we have occluded images.

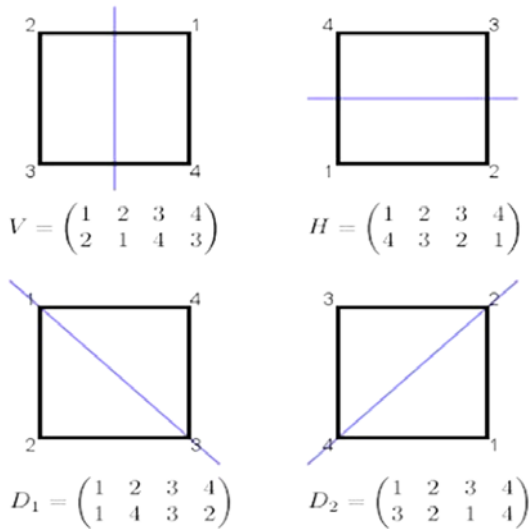

**Fig. 5.** Vertical, horizontal, and two diagonal reflections and their transformation matrices for facial key point detection.

In the Figure.5, **V**, **H** and **D$_i$** for **i** ∈ {1, 2} are as follows:

- **V** is flipped around vertical axis,
- **H** is flipped around horizontal axis,
- **D$_1$** is flipped around diagonal 1-3 and
- **D$_2$** is flipped around diagonal 2-4.

These matrixes show the reflection tips for the left and right of the brows, eyes and lips.

## 5 Algorithm

In the first step a region with the correct symmetry center of the image is selected and then by dihedral group filters, will find the rotated and flipped version of the current image and train the network with these original and converted (transformed) image versions. The major parts of this algorithm can be summarized in 3 steps:
1. Finding the facial image center with a verified algorithm after doing preprocessing on images,
2. Using Dihedral groups for finding accurate and precise rotated and flipped images,
3. Appling learning algorithm which presented the best results in facial expression researches such as NN and CNN in previous researches.

Here, we explain the general algorithm steps in details:

1. Doing preprocessing of the facial images such as de noising and detecting and cropping the face in the image as follows:
   a) Detecting the face area [35], [41]
   b) Crop eyebrow, eyes, and mouth region
   c) Finding Canny edge [10]
   d) Finding polygon over portions (Here Rectangle)
2. Finding Symmetry Points,
3. Finding the rotated and flipped image based on the dihedral rotation and flipped matrix in D$_4$ (rectangle).
4. We have 2 options here:
   a) Training the learning method such as neural network (NN) with all original and transformed images (rotated and flipped images).
   b) Using the transformation matrixes as filters in Convolutional Neural Network (CNN).
5. Testing with new data and using all data (old and new) for more training.

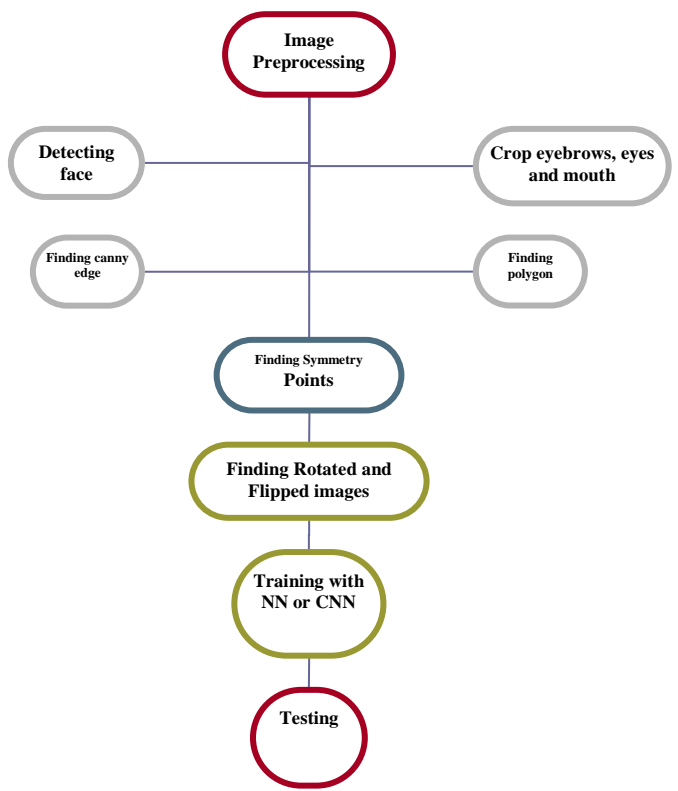

**Fig. 6.** Algorithm Flowchart.

In next research, we will do the steps on some static offline data for training, but the tests, will be in the real-time video data. Figure 6 shows these steps in the flowchart.

## 6 Conclusion and Future Work

In this article, we described how to use the dihedral group for symmetry detection in facial expression recognition and use it to recognize emotion. Based on the potential of the methods which it has some robust mathematical basis, theoretically we should get very fast and robust results. In the future work, we plan to apply the method with convolutional neural networks and use it for emotion recognition in video streaming data, especially when the data are corrupted or occluded [33], [34].